\documentclass[acmtog,nonacm,balance=false]{acmart}

\def\arxivversion{}

\newif\ifsubmission
\submissionfalse
\ifdefined\arxivversion\submissiontrue\fi
\ifdefined\arxivversion
\newcommand{\cvpronly}[1]{}
\newcommand{\arxivonly}[1]{#1}
\else
\newcommand{\cvpronly}[1]{#1}
\newcommand{\arxivonly}[1]{}
\fi
\newcommand{\suppref}[2]{#1}

\ifsubmission

\newcommand{\acc}[1]{}

\newcommand{\dcc}[1]{}
\newcommand{\mac}[1]{}
\newcommand{\maneesh}[1]{}
\newcommand{\naomi}[1]{}
\newcommand{\opc}[1]{}
\newcommand{\claude}[1]{}
\else

\newcommand{\acc}[1]{\textcolor{red}{\normalfont\small[Anthony: #1]}}

\newcommand{\dcc}[1]{\textcolor{red}{\normalfont\small[Danny: #1]}}
\newcommand{\mac}[1]{\textcolor{orange}{\normalfont\small[MA: #1]}}
\newcommand{\maneesh}[1]{\textcolor{orange}{\normalfont\small[MA: #1]}}
\newcommand{\naomi}[1]{\textcolor{blue}{\normalfont\small[Naomi: #1]}}
\newcommand{\opc}[1]{\textcolor{purple}{\normalfont\small[Or: #1]}}
\newcommand{\claude}[1]{\ \textcolor{claudegreen}{\normalfont\small[Claude: #1]}}
\fi
\newif\iftrackedits
\ifsubmission\trackeditsfalse\else\trackeditstrue\fi
\iftrackedits\usepackage[normalem]{ulem}\fi

\definecolor{claudegreen}{RGB}{0,140,60}
\definecolor{claudeedit}{RGB}{200,0,150}%

\providecommand{\Description}[1]{}

\usepackage{multirow}
\usepackage{algorithm}
\usepackage{algorithmic}
\usepackage{tikz}
\usetikzlibrary{positioning}
\usepackage{pgfplots}
\pgfplotsset{compat=1.18}

\definecolor{projectpagecolor}{RGB}{204,0,153}%

\makeatletter
\def\@parfont{\bfseries}
\makeatother

\begin{document}

\title[One Take, Many Cameras]{\textit{One Take, Many Cameras:}\texorpdfstring{\\}{ }Audio-Grounded Shot Synthesis for Dialogue Re-Cinematography}

\author{Anthony Chen\texorpdfstring{\textsuperscript{*}}{}}
\affiliation{\institution{Stanford University}\city{Stanford}\country{USA}}
\affiliation{\institution{LTX}\city{Jerusalem}\country{Israel}}
\author{Naomi Ken Korem}
\affiliation{\institution{LTX}\city{Jerusalem}\country{Israel}}
\author{Matan Ben-Yosef}
\affiliation{\institution{LTX}\city{Jerusalem}\country{Israel}}
\author{Tavi Halperin}
\affiliation{\institution{LTX}\city{Jerusalem}\country{Israel}}
\author{Urska Jelercic}
\affiliation{\institution{LTX}\city{Jerusalem}\country{Israel}}
\author{Ofir Bibi}
\affiliation{\institution{LTX}\city{Jerusalem}\country{Israel}}
\author{Daniel Cohen-Or}
\affiliation{\institution{Tel Aviv University}\city{Tel Aviv}\country{Israel}}
\author{Or Patashnik}
\affiliation{\institution{Tel Aviv University}\city{Tel Aviv}\country{Israel}}
\author{Maneesh Agrawala}
\affiliation{\institution{Stanford University}\city{Stanford}\country{USA}}
\renewcommand{\shortauthors}{Chen et al.}
\thanks{\textsuperscript{*}Work done in part during an internship at LTX. Currently a visiting researcher at Stanford University. Contact: \href{mailto:atchen@stanford.edu}{atchen@stanford.edu}}
\thanks{Images and videos are included for illustrative purposes only. Rights to any third-party material remain with their respective rights holders}
\authorsaddresses{}

\begin{abstract}
Creating a dialogue scene typically requires cutting between multiple views (e.g., close-ups, medium shots, and over-the-shoulder shots) of the actors. Yet capturing such coverage is costly and in many settings unavailable. We study \emph{dialogue re-cinematography}: synthesizing additional shots from a single wide-angle dialogue video. Existing video generators can synthesize new shots, but rely on visual cues alone, so when the source video barely shows the lips, the generated lip motion fails to match the speech. Our key observation is that the speech audio is view-invariant: it constrains lip motion in any target view. Building on this insight, we adapt a pretrained audio-visual diffusion model to generate the new shot from a target frame that specifies the view, a source video that contains the performance, and the speech audio that drives lip motion. In multi-person scenes, however, pretrained audio-visual models drive every visible face with the same speech, causing \emph{speech leakage}, where silent listeners mouth the speaker's words. We therefore introduce speaker-aware audio gating, a training-free module that lets the audio drive only the faces that speak in the target shot. On a cinematic paired-view benchmark spanning diverse viewpoints and framings, our method lowers perceptual error by 59\%, landmark error by 56\%, and speech leakage by 97\% relative to the strongest baseline, Wan~3.0. Our method also remains robust on in-the-wild scenes, highlighting its potential for film post-production. Project page: \href{https://recinematography.github.io/}{\textcolor{projectpagecolor}{https://recinematography.github.io/}}

\end{abstract}

\begin{teaserfigure}
  \centering
  \includegraphics[width=0.92\textwidth]{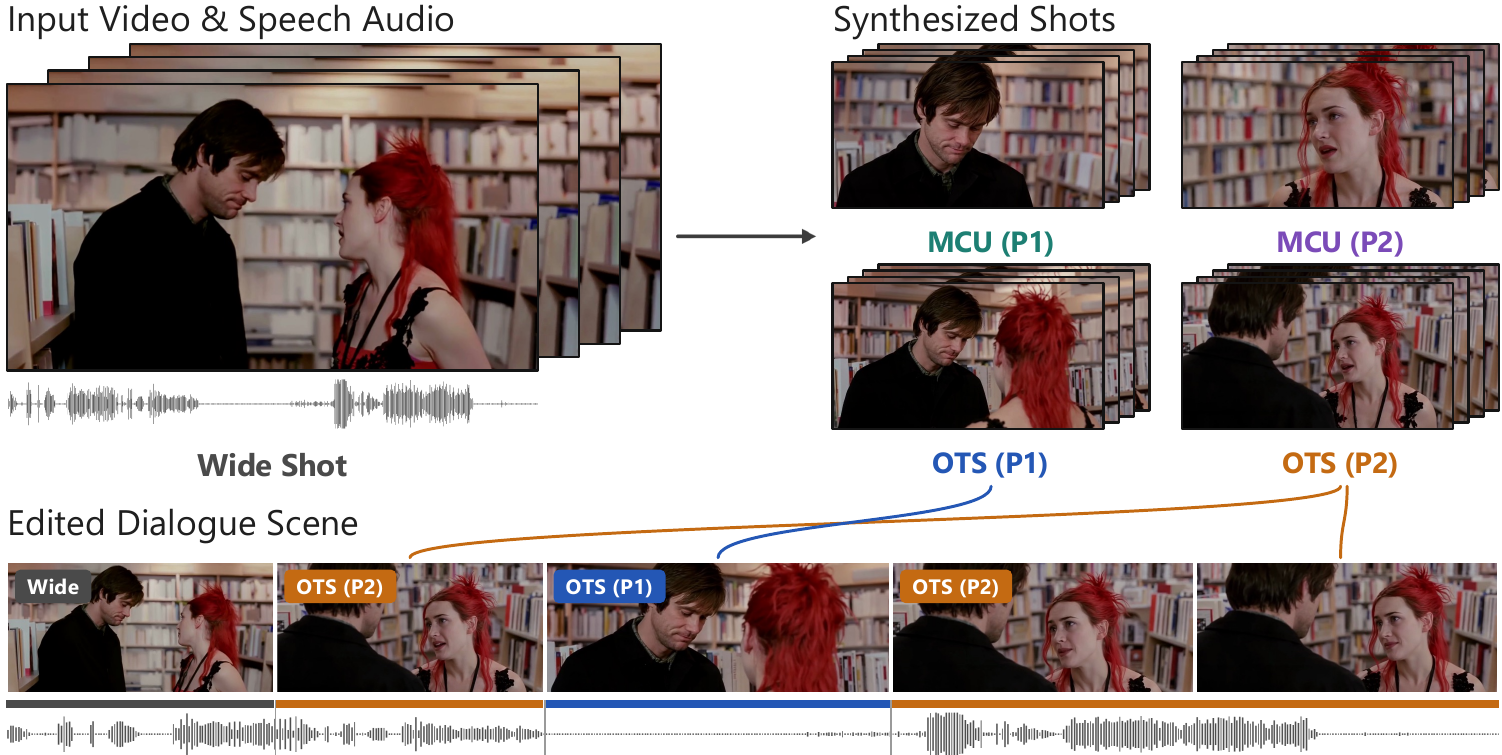}%
  \caption{From a single wide-shot dialogue video and its audio (upper left), our model synthesizes new cinematic shots of the same take, such as medium close-ups (MCU) and over-the-shoulder (OTS) views of each person (upper right). An editor can then cut between these shots as if the scene had been filmed with multiple cameras (bottom).}
  \label{fig:teaser}
\end{teaserfigure}

\maketitle

\section{Introduction}
\begin{figure*}[!t]
  \centering
  \includegraphics[width=\textwidth]{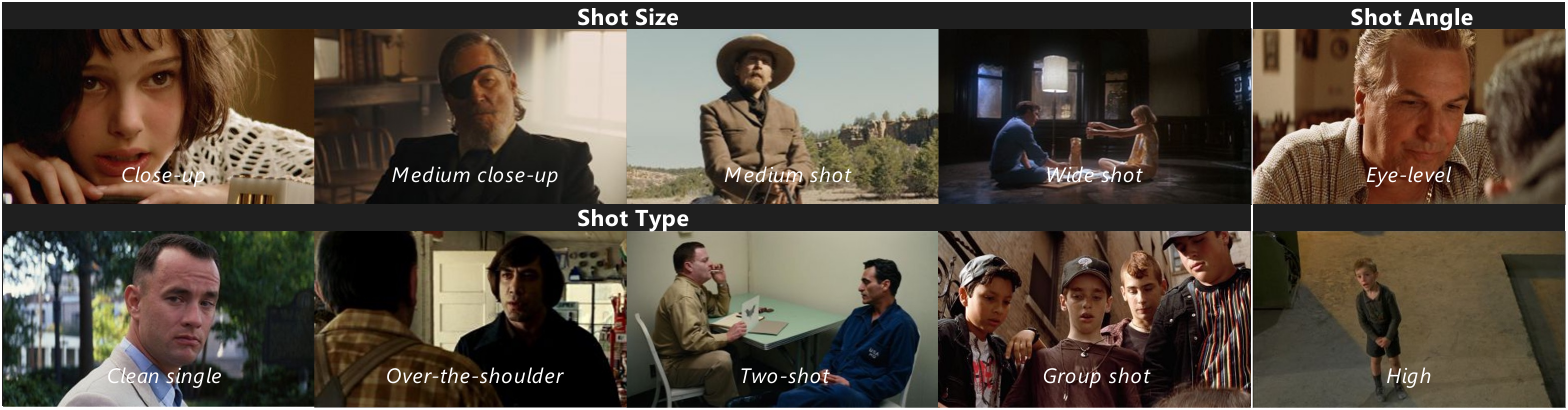}%
 \caption{Illustration of common cinematic vocabulary used in dialogue scenes.}
  \Description{A two-row grid of eight film frames from dialogue scenes, each labelled with a cinematic term. The upper row shows three shot sizes (close-up, medium shot, wide shot) and an eye-level example under the shot angle heading. The lower row shows three shot types (clean single, over-the-shoulder, group shot) and a high-angle example.}
  \label{fig:shot_vocabulary}
  \vspace{-6pt}%
\end{figure*}

\looseness=-1 Films often carry their story through conversation, and the dramatic force of a dialogue scene depends as much on editing as on performance. An editor can only cut between shots that were filmed, so filmmakers shoot each scene with \emph{coverage}---capturing a set of shots for the editor to choose from~\cite{Arijon1976, Murch2001}. For dialogue, typical coverage includes a wide master shot, close-ups and over-the-shoulder shots of the speaker, and reaction shots of the listeners (Figure~\ref{fig:shot_vocabulary}). Capturing these shots requires multiple camera positions, either with several synchronized cameras on a single take or over repeated takes. Both are costly and often infeasible for independent films, documentaries, and interviews. We therefore study \emph{dialogue re-cinematography}: synthesizing additional shots from a single wide-angle dialogue video as if the take had been filmed with multiple cameras (Figure~\ref{fig:teaser}).

\looseness=-1 Conventional re-cinematography methods simulate additional cameras with a crop window that pans, tilts, and zooms within the source video~\cite{Gleicher2008, Gandhi2014, ConvCut2019, EditIQ2025}. A crop, however, never moves the camera: a close-up only enlarges the source pixels, and an over-the-shoulder shot is impossible when the source video films the actors in profile. In contrast, recent video generation methods can move the camera by conditioning on the source video and a camera trajectory~\cite{ReCamMaster2025, ReCapture2025, TrajectoryCrafter2025, GEN3C2025, Vista4D2026, ReshootAnything2026, GoWithTheTrack2026}, synthesizing content that the source video never captured. However, when the source video shows an actor in profile, from behind, or too far away for the lips to be read clearly, these models hallucinate lip motions that fail to match the dialogue. Viewers are sensitive to mismatches between lips and speech~\cite{ITU1359, Vatakis2006, McGurk1976}, which makes these models impractical for dialogue re-cinematography.

\begin{figure*}[!t]
  \centering
  \includegraphics[width=\textwidth]{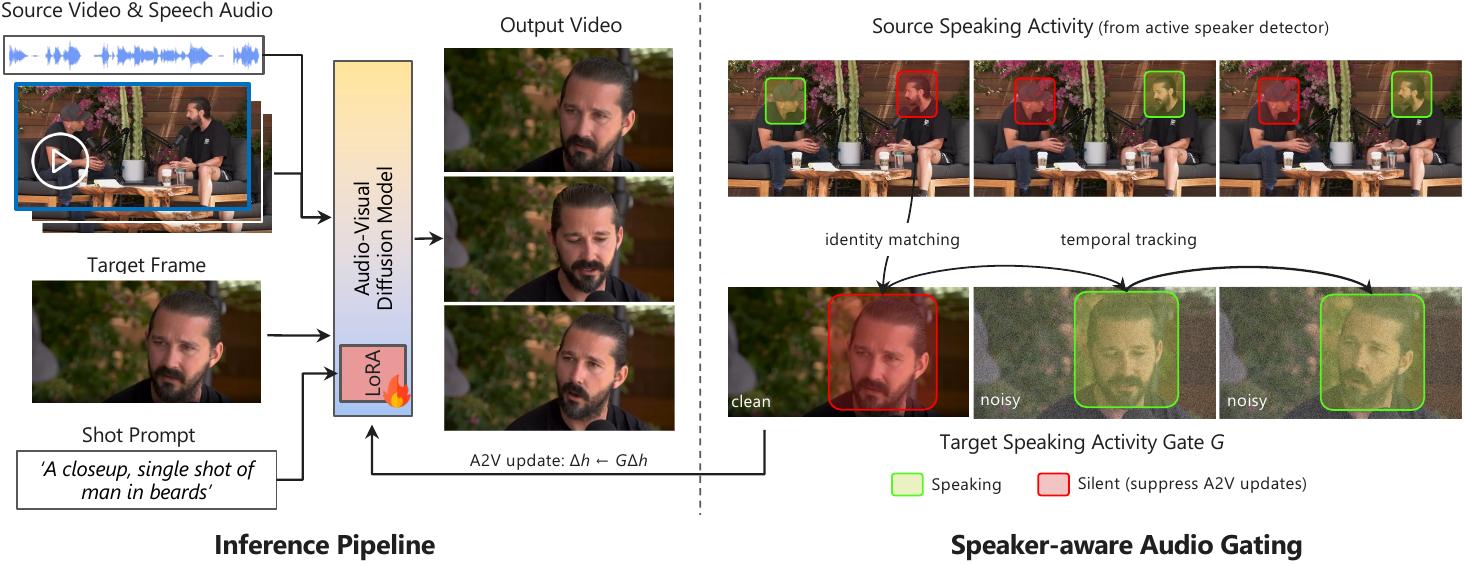}
  \caption{Method overview. From a source video with its speech audio, a target frame of the desired shot, and its shot prompt, our audio-visual diffusion model generates
the synchronized target-view video. \emph{Right:} speaker-aware audio gating. To prevent silent listeners from mouthing the speaker's words (speech leakage), an active-speaker detector labels each face in the source as speaking or silent. Cross-view attention correspondence matches each source face to the target frame,
cross-frame correspondence tracks it through the generated frames, and the gate $G$ suppresses audio-to-video updates on silent faces.}
  \Description{Left: source video and speech audio, a target frame, and a shot prompt feed an audio-visual diffusion model with LoRA, which outputs three frames of a close-up of a bearded man. Right: three source frames with speaking (green) and silent (red) boxes on two men; attention correspondence maps them to the target frames (one clean, two noisy), forming the gate G, which scales the audio-to-video updates of the diffusion model.}
  \label{fig:method_overview}
  \label{fig:pipeline}
  \label{fig:gating}
  \vspace{-6pt}%
\end{figure*}
Our key observation is that the audio can supply information that the pixels may lack. Speech is tightly coupled to lip motion~\cite{Yehia1998, Chandrasekaran2009} and is shared among all possible viewpoints. When an actor says ``map,'' the lips close on /m/, whether the camera faces the actor or sees only the back of the head. The audio is thus a view-invariant anchor for the lip motion in every new shot.

Building on this observation, we adapt a pretrained audio-visual diffusion model~\cite{LTX2_2026} to generate a new shot conditioned on the source video and its speech audio. The source video carries the performance, and the audio grounds the lip motion. Users specify the desired shot by providing its first frame, usually framed as one of the common cinematic shots (Figure~\ref{fig:shot_vocabulary}). An off-the-shelf image editing model can produce this frame from the source video's opening frame and a cinematic prompt, such as ``a close-up of the man.'' The filmmaker can thus preview several framings as still images and pick one before generation.

In multi-person scenes, grounding lip motion in speech has to be handled with care. Specifically, the speech must be bound to the person who utters it. Audio-visual video diffusion models typically inject the audio into every video token through cross-attention, resulting in \emph{speech leakage}, where silent listeners mouth another participant's words~\cite{polaczek2026attentiontriangleaudiovideomodels}.
We therefore introduce \emph{speaker-aware audio gating}, a training-free module that lets the audio drive only the faces of the active speakers. It first labels each face in the source video as speaking or silent with an off-the-shelf active-speaker detector. During denoising, it then transfers these labels to the corresponding faces in the new shot through the model's attention and masks the audio conditioning on the silent faces, all in a single generation pass.

We evaluate our approach on two benchmarks. A paired-view benchmark uses two cameras that captured the same take from different viewpoints. An in-the-wild benchmark covers diverse crowded scenes, each pairing a wide-shot video with a close-up frame. On the paired views, our method lowers perceptual error by 59\%, landmark error by 56\%, and speech leakage by 97\% relative to the strongest baseline, Wan~3.0~\cite{Wan3_2026}. On the in-the-wild clips, our model generalizes to crowded real-world scenes and is preferred over all baselines in 75\% of user-study responses.

\section{Related Work}

\paragraph{Computational Cinematography and Editing.}
Computational cinematography automates the camera decisions of film grammar~\cite{Arijon1976}. Such systems encode film idioms to place virtual cameras and cut between them, or plan shots from scene geometry, example clips, or on-screen constraints~\cite{He1996, ChristieCON2008, Galvane2018, Jiang2020, PulpMotion2026}. Computational editing applies these idioms to footage and selects shots and cut points across several cameras or takes for dialogue scenes~\cite{Berthouzoz2012, Arev2014, Leake2017}, narrated videos~\cite{QuickCut2016}, and interviews~\cite{ChunkyEdit2024}, and recent tools refine the timing of individual cuts~\cite{MicroEdits2026}. When only one camera is available, re-cinematography simulates the missing cameras by transforming the frames of the single recording~\cite{Gleicher2008}. These methods typically move pan-tilt-zoom crop windows within a wide-angle or 360-degree recording and edit between the resulting virtual shots~\cite{Gandhi2014, GAZED2020, ConvCut2019}. EditIQ~\cite{EditIQ2025}, for example, edits a static wide-angle dialogue recording by choosing among such virtual cameras based on the dialogue, the speakers, and visual saliency. A simulated camera, however, stays at the optical center of the real one. It can pan, tilt, and zoom, but it cannot move. A close-up therefore is a lower-resolution enlargement, and a shot from a new angle is out of reach. We instead synthesize the new views with generative models.

\paragraph{Novel-View Synthesis.}
Reconstruction-based methods fit radiance fields or Gaussians to a dynamic scene~\cite{NeRF2020, GaussianSplatting2023, FourDGS2024}, but reach high fidelity on faces and bodies only with many cameras~\cite{Lombardi2019, Isik2023} and degrade quickly from a single static view. Multi-shot video generation methods synthesize several shots of one scene~\cite{ShotAdapter2025, ShotDirector2026, HoloCine2025, MultiShotMaster2025}, but they generate the scene from a text script alone and cannot re-film an existing take. Generative reshooting methods re-film an existing take following an input camera trajectory with a video diffusion model~\cite{ReCamMaster2025, ReCapture2025, TrajectoryCrafter2025, GEN3C2025, Vista4D2026, ReshootAnything2026}. However, these methods condition on pixels alone, so they lack cues for lip motion when a face barely visible in the source fills the target view, for example when a profile wide shot is recut as a frontal close-up. The video prior then hallucinates articulation~\cite{HybridCinematography2026}. We instead build on an audio-visual foundation model so that the original speech drives articulation.

\paragraph{Reference-Guided Video Editing.}
Video editing models have developed from following text instructions~\cite{TuneAVideo2023, TokenFlow2024, InsV2V2024, VideoXQwen2026} to following visual references~\cite{UniVideo2026, KiwiEdit2026}. Reference images can guide object insertion~\cite{KiwiEdit2026, VINO2026, Bernini2026}, keyframe edit propagation~\cite{AnyV2V2024, I2VEdit2024, polaczek2026synclora}, and character animation and motion transfer~\cite{FOMM2019, AnimateAnyone2024, WanAnimate2025, SteadyDancer2025}. However, most of these editors are trained with references that share the same viewpoint as the source video. Given a frame from a completely different viewpoint, they tend to ignore it and fall back to the source view~\cite{MotionV2V2025}. Without audio guidance, they also fail to generate correct lip sync. Recent omni-reference foundation models, such as MiniMax~H3~\cite{MiniMaxH3_2026} and Wan~3.0~\cite{Wan3_2026}, take image, video, and audio references together. Applied to dialogue re-cinematography, they drive correct lip articulation from the audio but still show speech leakage, sometimes ignore the target frame, and hallucinate motion.
Multi-person talking-video models mitigate speech leakage by requiring a separate audio track for each person~\cite{MultiTalk2025, HunyuanVideoAvatar2025}. In contrast, our model is optimized to transfer the original performance to the target view, and our speaker-aware audio gating suppresses speech leakage from a single mixed audio track.

\section{Method}

Given a source dialogue video $V_s$, its audio track $A$, and a target frame $I_t$, we synthesize the requested target view while preserving the source performance (Figure~\ref{fig:pipeline}). The target frame sets the framing (Section~\ref{sec:cinematic_control}), and a pretrained audio-visual model, adapted with LoRA~\cite{LoRA2022}, takes the performance from the source video and the lip motion from the audio (Sections~\ref{sec:av_generation} and~\ref{sec:training}). At inference, speaker-aware gating keeps silent faces from being driven by another participant's speech (Section~\ref{sec:inference}).

\subsection{Cinematic Shot Control}
\label{sec:cinematic_control}

\looseness=-1 The new shot is specified by providing a target frame $I_t$, which will be the first frame in the generated shot. To create $I_t$, an off-the-shelf image editing model~\cite{InstructPix2Pix2023, QwenImage2025, GPTImage2_2026, NanoBananaPro2025, Seedream4_2025, HunyuanImage3_2025, FLUX2Klein2026} can edit the source video's first frame following a shot prompt in common cinematic vocabulary~\cite{ThompsonBowen2022} (Figure~\ref{fig:shot_vocabulary}), such as ``a medium close-up of the woman, at eye level.'' Comparing several editing models (supplementary \suppref{Table~\ref{tab:image_generation_benchmark}}{Section~C}), we find that GPT-Image~2 performs best, so we use it to generate the target frames in Figure~\ref{fig:qual_shots_more}. In our quantitative evaluation, we use the ground-truth target frame (Section~\ref{sec:exp_setup}).

\subsection{Audio-Visual In-Context Conditioning}
\label{sec:av_generation}

\begin{figure}[!t]
  \centering
  \includegraphics[width=0.9\columnwidth]{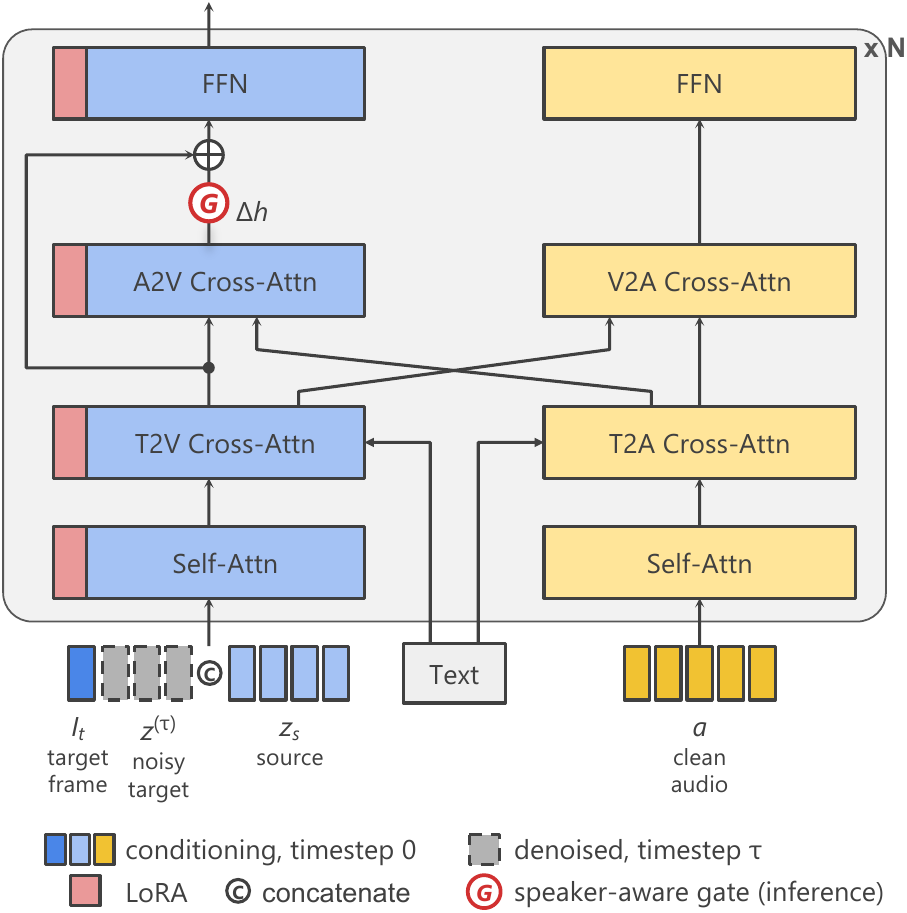}
  \caption{LTX-2.3~\cite{LTX2_2026} model architecture. We inject conditioning tokens in context and train LoRA adapters on the video stream. Speaker-aware gating is applied to the A2V residuals.}
  \label{fig:method_arc}
  \vspace{-6pt}
\end{figure}

We build on LTX-2.3~\cite{LTX2_2026}, a pretrained audio-visual diffusion transformer with separate video and audio streams (Figure~\ref{fig:method_arc}). Video and audio VAEs encode the two modalities into latent tokens, and the diffusion model combines them through within-modality self-attention, text cross-attention, and bidirectional audio-video cross-attention. Trained jointly on audio and video, the model generates videos that synchronize lip motion with speech. Here we adapt the model to accept audio-visual conditioning with an in-context mechanism~\cite{InContextLoRA2024, benyosef2026avcontrol, VACE2025, JustDubIt2026, polaczek2026synclora}.

\paragraph{Visual Conditioning.}
We encode the target frame $I_t$ with the video VAE $\mathcal{E}_v$ as the clean first-frame latent of the target video and initialize the remaining target tokens with noise. The source video is encoded into clean tokens $z_s=\mathcal{E}_v(V_s)$ and concatenated with the target sequence. In LTX-2.3, each token has its own timestep, allowing us to set all conditioning tokens to timestep $0$, while the target tokens follow the denoising schedule. This way, at every denoising step, the noisy target tokens attend to the clean source tokens.

\paragraph{Audio Conditioning.}
We encode the audio waveform via the audio VAE $\mathcal{E}_a$ into audio latents $a=\mathcal{E}_a(A)$, feed them through the audio stream, and hold every audio token clean at timestep $0$, so the clean speech conditions the video tokens.
\looseness=-1 We generate only the video stream and pair it with the original audio, so generated shots can be cut against that audio without post-processing.

\subsection{Training}
\label{sec:training}
\begin{figure*}[t]
  \centering
  \includegraphics[width=\textwidth]{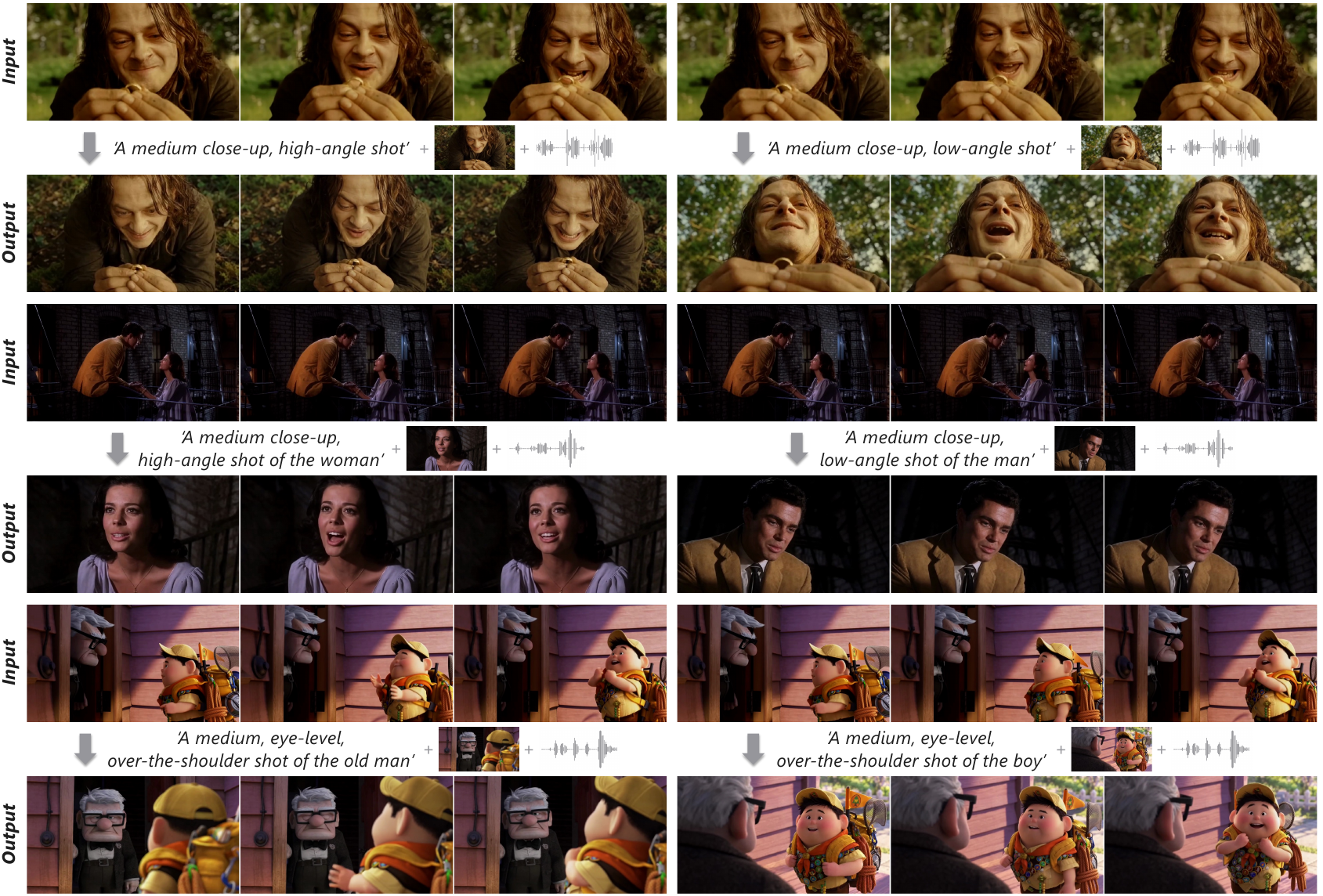}
  \caption{Results on dialogue scenes from films. Target frames are generated by GPT-Image~2 from the source frame and the shot prompt. See the videos on the project page for better visualization.}
  \label{fig:qual_shots_more}
  \vspace{-6pt}%
\end{figure*}

\paragraph{Data Construction.}
We use open-source multi-view datasets that cover diverse human activity, from full-body motion (CMU Panoptic~\cite{CMUPanoptic2017}, PKU-DyMVHumans~\cite{PKUDyMVHumans2024}, SynCamVideo~\cite{SynCamVideo2025}) to single-speaker talking faces (MEAD~\cite{MEAD2020}) and multi-person conversations (Replay~\cite{Replay2023}, DnD Group Gesture~\cite{ConvoFusion2024}). For each training pair, we randomly sample two synchronized videos of a scene from different viewpoints and use one as the source $V_s$ and the other as the target $V_t$. For robustness to diverse framings, we crop the target video to different shot sizes and shot types, using estimated body keypoints to place each crop (examples in supplementary \suppref{Figure~\ref{fig:data_construction}}{Section~A}). The target frame $I_t$ is the first frame of the target video. The caption states the shot size, type, and angle, followed by a scene and motion description from Qwen3-VL~\cite{Qwen3VL2025}.

\paragraph{Training Objective.}
We initialize LoRA adapters on the video stream
and train with the flow-matching objective~\cite{Lipman2023}.
Let $z_{\mathrm{tgt}}$ be the target video latents, excluding the target frame.
\looseness=-1 For noise $\epsilon \sim \mathcal{N}(0,I)$ and timestep $\tau$, we form $z^{(\tau)} = (1-\tau)z_{\mathrm{tgt}} + \tau\epsilon$ and train $v_\theta$ to predict the velocity $\epsilon-z_{\mathrm{tgt}}$:
\begin{equation}
\mathcal{L}=\mathbb{E}\big[\|v_\theta(z^{(\tau)},c,\tau)-(\epsilon-z_{\mathrm{tgt}})\|^2\big],
\end{equation}
\looseness=-1 where $c$ denotes the clean conditioning latents: the source-video tokens and the target frame, and additionally the audio tokens for clips recorded with speech. The loss is computed only on the target video tokens.

\subsection{Inference with Speaker-Aware Audio Gating}
\label{sec:inference}
\label{sec:speaker_gating}

Pretrained audio-visual models produce accurate lip motion through audio-to-video cross-attention. This attention, however, updates every video token from the same audio tokens without knowing which face is speaking, so silent listeners also mouth the speaker's words~\cite{polaczek2026attentiontriangleaudiovideomodels}. To prevent this speech leakage, we add a training-free module at inference that suppresses audio-to-video updates on silent faces. We detect who speaks in the source and map the labels to the target view through the model's attention (Figure~\ref{fig:gating}).

\paragraph{Speaker Detection.}
Before generation, we run LightASD~\cite{LightASD2023}, an off-the-shelf active-speaker detector, on the source video $V_s$ and audio $A$ to obtain a face track and per-frame speaking labels for each source participant $p$, downsampled to latent time as $y_{p,t}\in\{0,1\}$.

\paragraph{Identity Matching.}
Source and target video tokens are concatenated at the input (Section~\ref{sec:av_generation}), so they attend to each other at every visual self-attention layer. We use the clean first target frame (the encoded $I_t$) as the reference. During the first $s_g$ denoising steps, audio-to-video updates are disabled to avoid speech leakage before the gate exists. From the reference frame's attention to the first source frame, averaged over heads, selected blocks, and these steps, we build a correspondence map $W_{ct}$, normalized in both directions so that a source--target pair scores high only when the two locations match each other. Let $\mathcal{S}_{p,a}$ contain the source cells of facial patch $a$ of participant $p$. The correspondence mass that target location $i$ sends to this patch is
\begin{equation}
R_{p,a}(i)=\sum_{j\in\mathcal{S}_{p,a}}W_{ct}(i,j).
\label{eq:face_localization}
\end{equation}
Placing each facial patch at its peak target location locates the face, and the located face keeps the identity and speaking labels $y_{p,t}$ of its source participant. Participants with weak or ambiguous correspondence, such as those off-screen, are left ungated, so an unreliable match cannot silence a speaker. Both identity matching and the tracking below read the model's attention maps, so no separate face detector or re-identification model is needed.

\paragraph{Face Tracking.}
Only the first target frame is clean while the gate is built, and the remaining frames are still noisy latents. Following DiffTrack~\cite{Nam2025DiffTrack}, we track each located face through the target frames with bidirectional cross-frame attention correspondences, which already encode coarse layout and temporal correspondence in the early, noisy denoising steps~\cite{Nam2025DiffTrack, Choi2022P2}. A soft mask $M_{p,t}(i)\in[0,1]$ centered on the smoothed track, together with $y_{p,t}$, forms the speaking-activity mask of the target video. On held-out multi-person clips, attention-based identity matching is 96.3\% accurate and tracking deviates by 5.3 latent cells on average.

\begin{table*}[!t]
  \centering
  \caption{Quantitative results on the paired-view benchmark. Source copy: share of clips whose output looks more like the source view than the target view. $\uparrow$/$\downarrow$: higher/lower is better. \textbf{Bold}/\underline{underline}: best/second best.}
  \label{tab:paired_view}
  \renewcommand{\arraystretch}{1.05}
  \footnotesize
  \setlength{\tabcolsep}{2.3pt}
  \begin{tabular*}{\textwidth}{@{\extracolsep{\fill}}l ccc ccc cc ccc c@{}}
    \toprule
    & \multicolumn{3}{c}{Appearance Metrics}
    & \multicolumn{3}{c}{Facial Metrics}
    & \multicolumn{2}{c}{Body Metrics}
    & \multicolumn{3}{c}{Dialogue Metrics}
    & \multicolumn{1}{c}{View} \\
    \cmidrule(lr){2-4}\cmidrule(lr){5-7}\cmidrule(lr){8-9}\cmidrule(lr){10-12}\cmidrule(l){13-13}
    Method
    & LPIPS\,$\downarrow$ & MSE\,{\tiny$\times$1000}\,$\downarrow$ & ArcFace\,$\uparrow$
    & LMD\,$\downarrow$ & AED\,$\downarrow$ & Gaze\,$^\circ$\,$\downarrow$
    & PCK\,\%\,$\uparrow$ & Vel.\,$\downarrow$
    & mAP\,\%\,$\uparrow$ & Leakage\,\%\,$\downarrow$ & Sync-C\,$\uparrow$ & Source copy\,\%\,$\downarrow$ \\
    \midrule
    MiniMax H3~\cite{MiniMaxH3_2026}
      & 0.290 & 34.69 & 0.785 & 0.174 & 0.127 & 5.69 & 29.3 & 0.138 & 62.4 & 47.3 & \underline{5.29} & 26.4 \\
    Wan 3.0~\cite{Wan3_2026}
      & \underline{0.235} & \underline{24.83} & \textbf{0.821} & \underline{0.142} & \underline{0.098} & \underline{4.51} & \underline{41.8} & \underline{0.107} & \underline{85.8} & \underline{17.5} & 5.07 & \underline{6.8} \\
    Ours
      & \textbf{0.097} & \textbf{7.48} & \underline{0.800} & \textbf{0.062} & \textbf{0.090} & \textbf{4.34} & \textbf{53.4} & \textbf{0.083} & \textbf{94.8} & \textbf{0.6} & \textbf{5.47} & \textbf{0.0} \\
    \bottomrule
  \end{tabular*}
  \vspace{-6pt}%
\end{table*}
\begin{figure*}[!t]
  \centering
  \begin{tikzpicture}
  \node[anchor=south west,inner sep=0] (img) at (0,0) {\includegraphics[width=\textwidth,pagebox=cropbox]{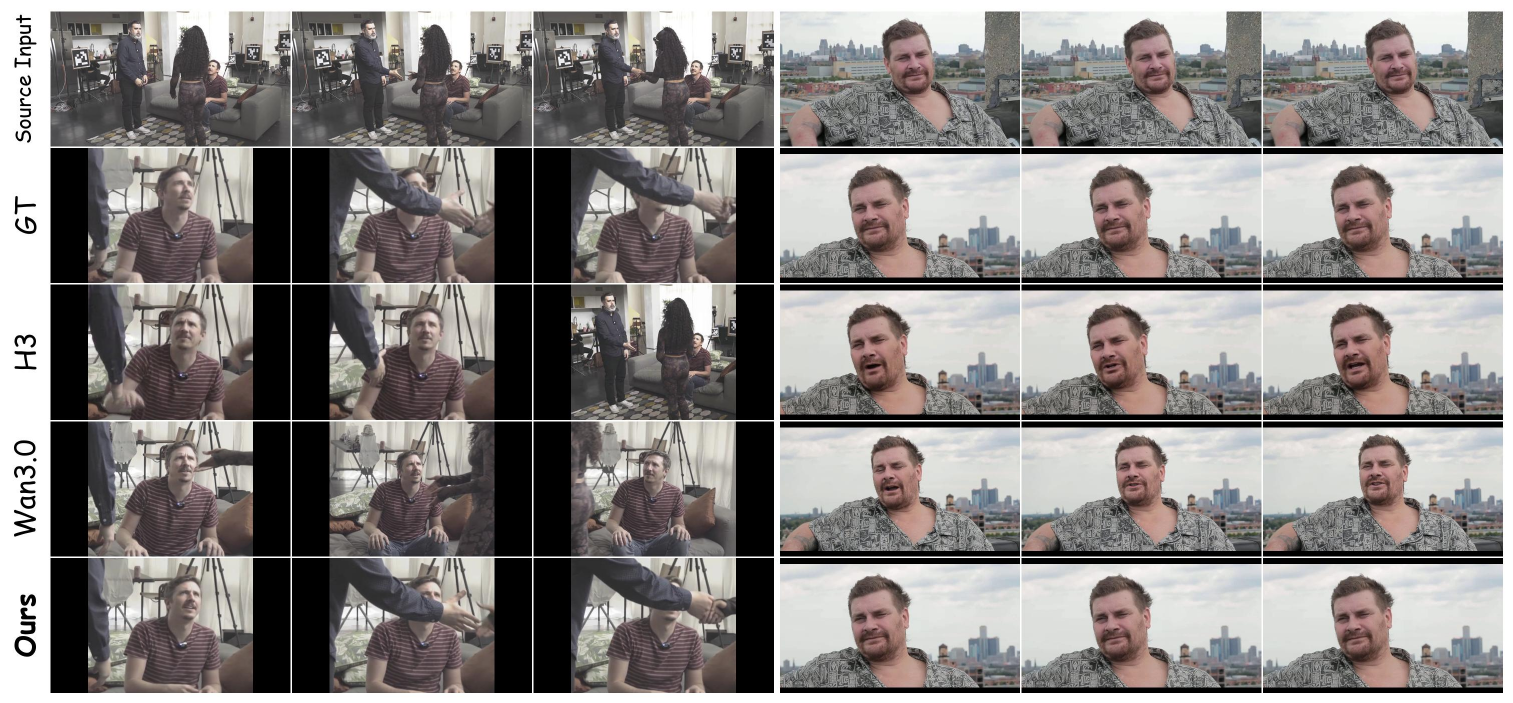}};
  \begin{scope}[x={(img.south east)},y={(img.north west)},failmark/.style={draw=red,dashed,line width=1.1pt}]
    \draw[failmark] (0.4317,0.5013) ellipse (0.0771 and 0.0933);
    \draw[failmark] (0.1437,0.3480) ellipse (0.0200 and 0.0251);
    \draw[failmark] (0.3175,0.2915) ellipse (0.0175 and 0.0825);
    \draw[failmark] (0.3858,0.3076) ellipse (0.0133 and 0.0897);
    \draw[failmark] (0.5771,0.4762) ellipse (0.0175 and 0.0233);
    \draw[failmark] (0.7375,0.4735) ellipse (0.0175 and 0.0233);
    \draw[failmark] (0.8979,0.4762) ellipse (0.0175 and 0.0233);
    \draw[failmark] (0.5883,0.3094) ellipse (0.0175 and 0.0233);
    \draw[failmark] (0.7488,0.3085) ellipse (0.0175 and 0.0233);
    \draw[failmark] (0.9092,0.3103) ellipse (0.0175 and 0.0233);
  \end{scope}
  \end{tikzpicture}
  \caption{Qualitative comparison on the paired-view benchmark. The baselines fail to transfer the source motion (left), show speech leakage (right), and miss the requested shot (both), marked by dashed circles. Ours achieves the strongest motion and dialogue consistency.}
  \Description{Two examples side by side, each with five rows of three frames: the source input, the ground-truth target view, H3, Wan 3.0, and ours. The left example is a wide indoor shot of two people greeting a seated man, with a close target view of the seated man as another person reaches toward him; the right example is a man in a patterned shirt outdoors in front of a city skyline.}
  \label{fig:paired_qual}
  \vspace{-6pt}%
\end{figure*}

\paragraph{Audio Gating.}
In LTX-2.3, each audio-to-video (A2V) cross-attention block adds an update $\Delta h_t(i)$ to the target video token at latent frame $t$ and location $i$. We replace it with $G_t(i)\,\Delta h_t(i)$, where
\begin{equation}
G_t(i)=\min_{p\in\mathcal{V}}\left[1-\left(1-y_{p,t}\right)M_{p,t}(i)\right]
\label{eq:speaker_gate}
\end{equation}
\looseness=-1 and $\mathcal{V}$ contains the located participants. The gate acts as a soft spatial mask: it is $0$ on silent faces and $1$ elsewhere, including active speakers and the background, since audio also drives non-speech content such as reactions to sounds. Only the A2V blocks are gated on target tokens. The layers that transfer body motion and listener reactions are not. Rough facial appearance is set in early denoising steps~\cite{Choi2022P2}, so we build the gate in the first $s_g$ steps and apply it in the remaining steps, where lip motion is mainly shaped~\cite{LipForcing2026}. Supplementary Section~\suppref{\ref{sec:app_gating}}{A} gives the full algorithm.

\section{Experiments}

\subsection{Evaluation Setup}
\label{sec:exp_setup}
\paragraph{Paired-View Benchmark.}
We first sample 100 cross-camera pairs from Replay~\cite{Replay2023}, using scenes held out from training, and crop the target video to cinematic framings such as close-ups and over-the-shoulder shots, keeping the source video wide. To increase benchmark scene diversity, we additionally collect 196 pairs of multi-camera footage from real productions on the internet. In each pair, the first frame of the second camera serves as the target frame and the rest as ground truth.%

\paragraph{In-the-Wild Benchmark.}
To evaluate our method on real-world cinematic scenes with crowded conversations, for which paired multi-camera footage is not publicly available, we collect 98 edited dialogue videos from the internet containing transitions between close shots and wider shots. For each pair, the wide-shot clip serves as the source video, while a frame from the corresponding close shot provides target-frame conditioning. The model then generates the wide-shot segment from the close-shot viewpoint. Since no ground-truth target video is available, we primarily evaluate speaking activity. Supplementary Figure~\ref{fig:supp_inthewild_construction} shows how the pairs are built.

\paragraph{Baselines.}
\looseness=-1 We compare with state-of-the-art reference-guided video generation models, the open-source MiniMax~H3~\cite{MiniMaxH3_2026} and the closed-source Wan~3.0~\cite{Wan3_2026}\footnote{Seedance~\cite{Seedance2_2026} and Gemini Omni~\cite{GeminiOmni2026} reject our inputs under their content policies and are therefore not listed as baselines.}. They receive the same inputs (i.e., video, audio, and frame) as our model during evaluation. Camera-controlled models (e.g., GEN3C~\cite{GEN3C2025}, Vista4D~\cite{Vista4D2026}) and reference-guided editors (e.g., VINO~\cite{VINO2026}, Bernini-R~\cite{Bernini2026}) produce strong artifacts or stay at the source framing in our task, so we show their results only qualitatively (supplementary \suppref{Figure~\ref{fig:inthewild_camctrl}}{Section~D}).

\begin{table}[t]
  {%
    \renewenvironment{table}[1][]{\captionsetup{type=table}}{\par\vspace{14pt}}%
    \begin{table}[t]
  \centering
  \caption{Quantitative results on the in-the-wild benchmark. User-study columns report the percentage of responses favoring each method for speaker attribution (Spk.), motion and lip sync (Sync), and overall preference (All).}
  \vspace{-7pt}
  \label{tab:inthewild_frame}
  \footnotesize
  \setlength{\tabcolsep}{2.5pt}
  \renewcommand{\arraystretch}{1.05}
  \begin{tabular*}{\columnwidth}{@{\extracolsep{\fill}}l cc ccc@{}}
    \toprule
    & \multicolumn{2}{c}{Dialogue Metrics}
    & \multicolumn{3}{c}{User Study} \\
    \cmidrule(lr){2-3}\cmidrule(l){4-6}
    Method & mAP\,$\uparrow$ & Leakage\,\%\,$\downarrow$ & Spk.\,$\uparrow$ & Sync\,$\uparrow$ & All\,$\uparrow$ \\
    \midrule
    MiniMax H3~\cite{MiniMaxH3_2026} & 65.5 & \underline{21.6} & 13.4 & 13.6 & 12.2 \\
    Wan 3.0~\cite{Wan3_2026} & \underline{78.1} & 27.5 & \underline{17.0} & \underline{14.8} & \underline{13.1} \\
    Ours & \textbf{90.3} & \textbf{11.9} & \textbf{69.6} & \textbf{71.6} & \textbf{74.7} \\
    \bottomrule
  \end{tabular*}
\end{table}
    \begin{table}[t]
  \centering
  \caption{Quantitative ablation on audio conditioning. Removing the audio input at inference desynchronizes lip motion. $^*$: computed on the mouth region.}
  \label{tab:ablation_audio}
  \footnotesize
  \setlength{\tabcolsep}{1pt}
  \renewcommand{\arraystretch}{1.1}
  \begin{tabular*}{\columnwidth}{@{\extracolsep{\fill}}lcccccc@{}}
    \toprule
    & \multicolumn{4}{c}{Paired-view} & \multicolumn{2}{c}{In-the-wild} \\
    \cmidrule(lr){2-5}\cmidrule(l){6-7}
    Variant & Sync-C\,$\uparrow$ & Sync-D\,$\downarrow$ & LPIPS$^*$\,$\downarrow$ & MSE$^*$\,{\tiny$\times$1000}\,$\downarrow$ & Sync-C\,$\uparrow$ & Sync-D\,$\downarrow$ \\
    \midrule
    w/ audio & \textbf{5.01} & \textbf{8.53} & \textbf{0.170} & \textbf{15.6} & \textbf{5.68} & \textbf{8.20} \\
    w/o audio & 3.70 & 9.74 & 0.172 & 15.8 & 3.35 & 10.45 \\
    \bottomrule
  \end{tabular*}
  \vspace{-7pt}%
\end{table}

\begin{table}[!t]
  \centering
  \caption{Quantitative ablation on speaker-aware audio gating over $15$ multi-person in-the-wild clips.}
  \label{tab:gating_ablation}
  \footnotesize
  \setlength{\tabcolsep}{3pt}
  \renewcommand{\arraystretch}{1.1}
  \begin{tabular*}{\columnwidth}{@{\extracolsep{\fill}}lccc@{}}
    \toprule
    Variant & Dialogue mAP\,$\uparrow$ & Speech leakage\,\%\,$\downarrow$ & Sync-C\,$\uparrow$ \\
    \midrule
    w/ gating & \textbf{86.4} & \textbf{5.0} & 5.18 \\
    w/o gating & 79.6 & 9.7 & \textbf{5.31} \\
    \bottomrule
  \end{tabular*}
  \vspace{-7pt}%
\end{table}
\begin{table}[t]
  \centering
  \caption{Quantitative ablation on crop augmentation.}
  \label{tab:ablation_training}
  \footnotesize
  \setlength{\tabcolsep}{1.5pt}
  \renewcommand{\arraystretch}{1.1}
  \begin{tabular*}{\columnwidth}{@{\extracolsep{\fill}}lccccc@{}}
    \toprule
    Variant & MSE\,{\tiny$\times$1000}\,$\downarrow$ & LMD\,$\downarrow$ & AED\,$\downarrow$ & PCK\,$\uparrow$ & Dialogue mAP\,$\uparrow$ \\
    \midrule
    Ours & \textbf{7.48} & \textbf{0.062} & \textbf{0.090} & \textbf{53.4} & \textbf{94.8} \\
    w/o crop aug. & 9.62 & 0.089 & 0.100 & 52.2 & 91.7 \\
    \bottomrule
  \end{tabular*}
  \vspace{-7pt}%
\end{table}
  }%
  \vspace{-14pt}
\end{table}

\paragraph{Evaluation Metrics.}
\label{sec:metrics}

We measure appearance against the ground-truth target view with LPIPS~\cite{LPIPS2018}, MSE, and ArcFace~\cite{ArcFace2019} identity similarity. We measure facial performance with landmark distance (LMD)~\cite{LMD_Chen2018}, DECA~\cite{DECA2021} expression error (AED), and OpenFace~2~\cite{OpenFace2_2018} gaze error, and body performance with PCK~\cite{PCK_YangRamanan2013} and joint-velocity error on RTMPose~\cite{RTMPose2023} keypoints. For speaker attribution, TalkNet~\cite{TalkNet2021}, which is separate from the detector used for gating, provides reference speaking labels and per-face scores on the output. Dialogue mAP is the mean average precision over speaking and non-speaking activity. Speech leakage measures the share of frames in which a face that is silent in the source speaks in the target output. SyncNet~\cite{SyncNet2016} confidence (Sync-C) and distance (Sync-D) measure lip sync. Source copy measures how often the output copies the source view, judged by SSIM~\cite{SSIM2004}.

\subsection{Results}
\label{sec:exp_results}
\label{sec:exp_quantitative}
We first show results on film dialogue scenes in Figure~\ref{fig:qual_shots_more}. Our model generalizes across diverse shots and to 3D cartoons (more examples in supplementary \suppref{Figure~\ref{fig:qual_shots_more_supp}}{Section~D}). Since lip sync and speech leakage are best judged in motion, we encourage viewing the videos on the project page.
\begin{figure}[t]
  \centering
  \includegraphics[width=\columnwidth,pagebox=cropbox,trim=0 265.3 0 0,clip]{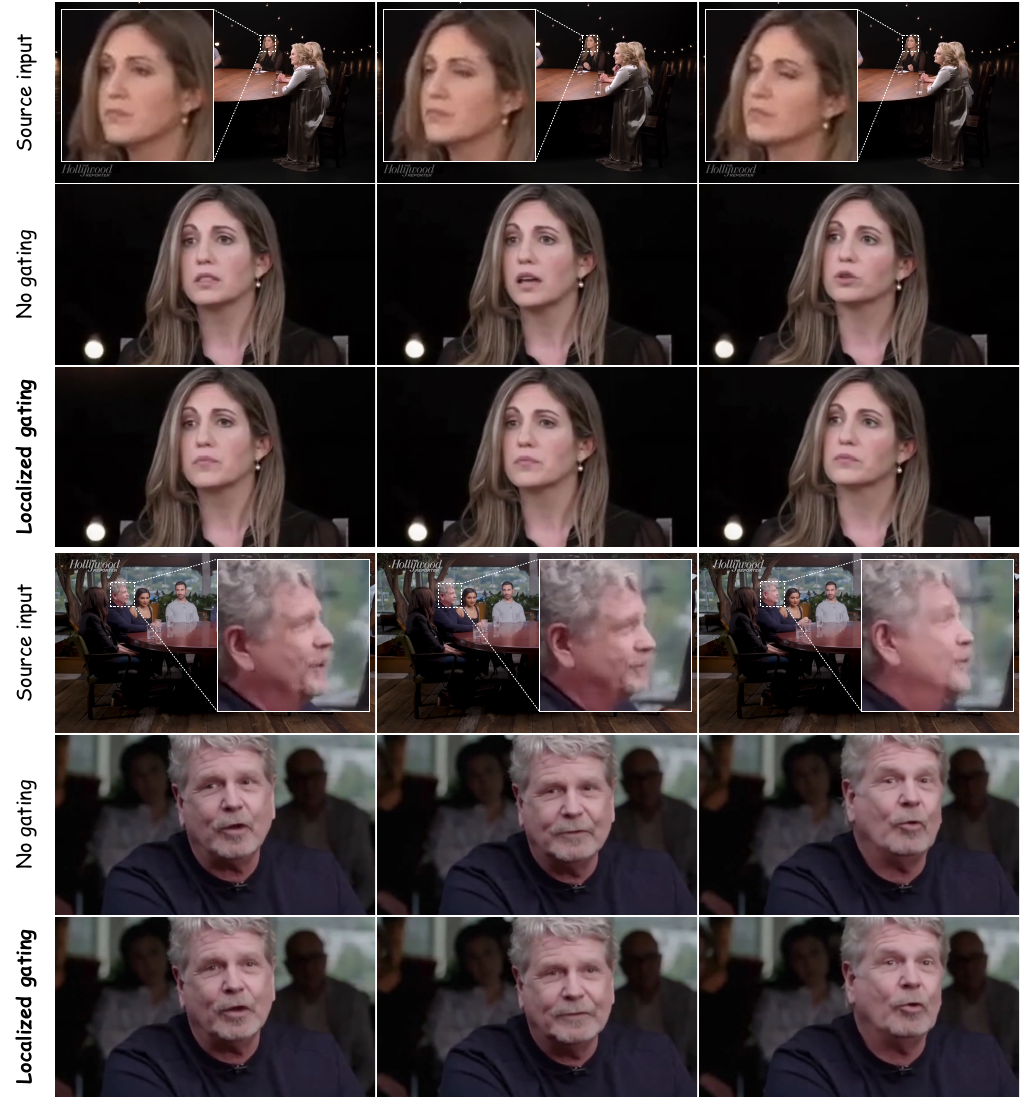}
  \caption{Ablation on speaker-aware audio gating. Without gating, the silent listener mouths the speaker's words. With gating, her mouth stays closed.}

  \Description{Two examples stacked vertically, each with three rows of three frames. In each example, the top row is the wide source input with a dashed box around the target subject and an enlarged inset of that subject, the middle row is the result without gating, and the bottom row is the result with gating. The first example is a woman with long hair at a round table; the second is an older man with grey hair at a panel table.}
  \label{fig:ablation_gating}
  \vspace{-7pt}%
\end{figure}

\paragraph{Paired-View Comparison.}
\label{sec:exp_paired}
\looseness=-1 Table~\ref{tab:paired_view} reports results on the paired-view benchmark, where each output is evaluated against the ground-truth target video. Our method outperforms both baselines on most metrics: the large gaps in LPIPS and MSE show a much closer match to the ground truth, and the facial, body, and dialogue metrics show that the captured performance is better preserved. On ArcFace similarity, our method stays competitive with the closed-source Wan~3.0, a commercial model built for reference identity consistency~\cite{Wan3_2026}, whereas our adapters are trained on only $9$k video pairs. Our method also has the lowest speech leakage together with the highest dialogue mAP, so it keeps listeners silent without muting the speakers. Unlike the baselines, which sometimes fall back to the source view, our method follows the requested shot in every clip.

Figure~\ref{fig:paired_qual} shows a qualitative comparison. On the left, neither baseline reproduces the handshake from the source video. On the right, both baselines move the listener's mouth with the off-screen interviewee's words. They also drift from the requested shot: H3 falls back to the wide framing (left), and Wan~3.0 misses the target framing at the start (right). Our method stays synchronized with the source and holds the framing.

\paragraph{In-the-Wild Comparison and Human Evaluation.}
\label{sec:exp_inthewild}
As shown in Table~\ref{tab:inthewild_frame}, our method achieves the highest dialogue mAP and the lowest speech leakage, showing that its speaker attribution stays robust in crowded real-world scenes. The baselines show more speech leakage and drift from the source gestures (visual examples in supplementary \suppref{Figure~\ref{fig:inthewild_camctrl}}{Section~D}). 

We further compare with the baselines in a user study with $22$ participants, yielding $528$ votes. For each pair of results, participants choose the better one on speaker attribution (whether the right person speaks while the listeners stay silent), motion and lip sync, and overall preference. Our method is preferred over both baselines on all three criteria (Table~\ref{tab:inthewild_frame}). %

\begin{figure}[t]
  \centering
  \includegraphics[width=\columnwidth]{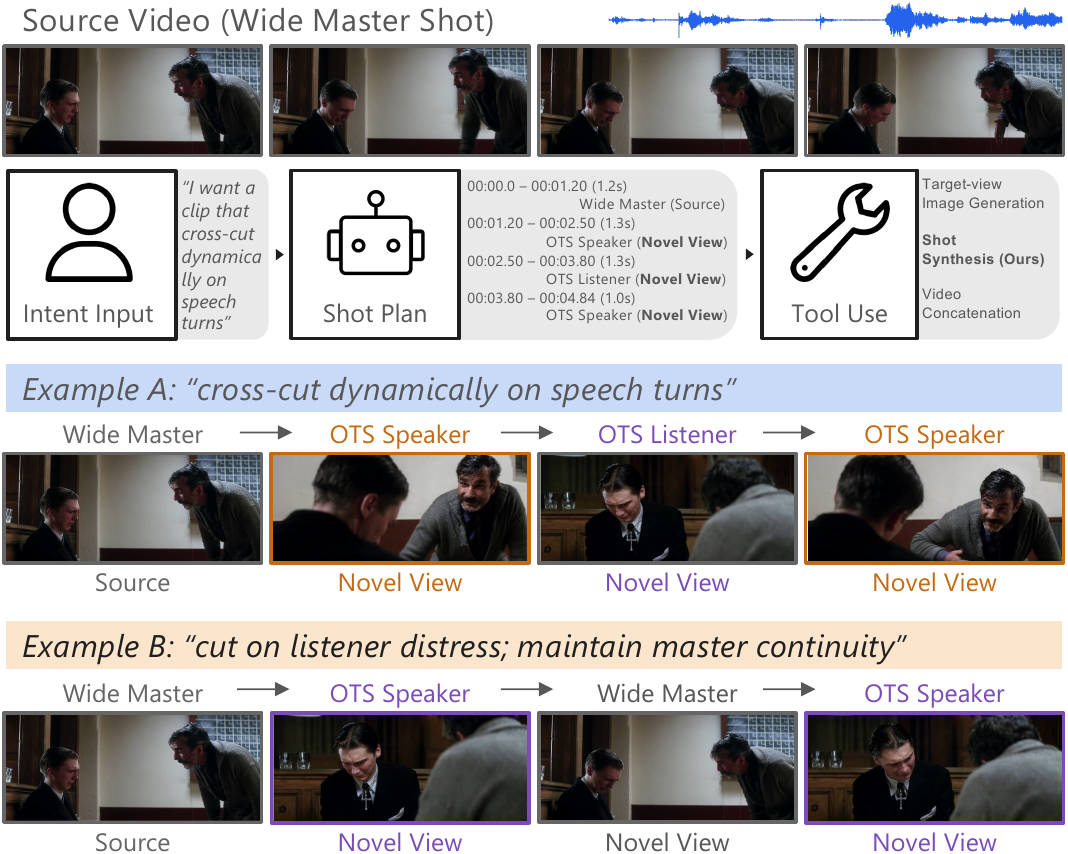}
  \caption{Agentic editing with generative coverage. An editing agent plans coverage from a style prompt and calls our method to generate shots the camera did not capture. Two contrasting prompts produce two different cuts of one wide master take.}
  \label{fig:autocine_demo}
\end{figure}

\begin{figure}[t]
  \centering
  \includegraphics[width=\columnwidth]{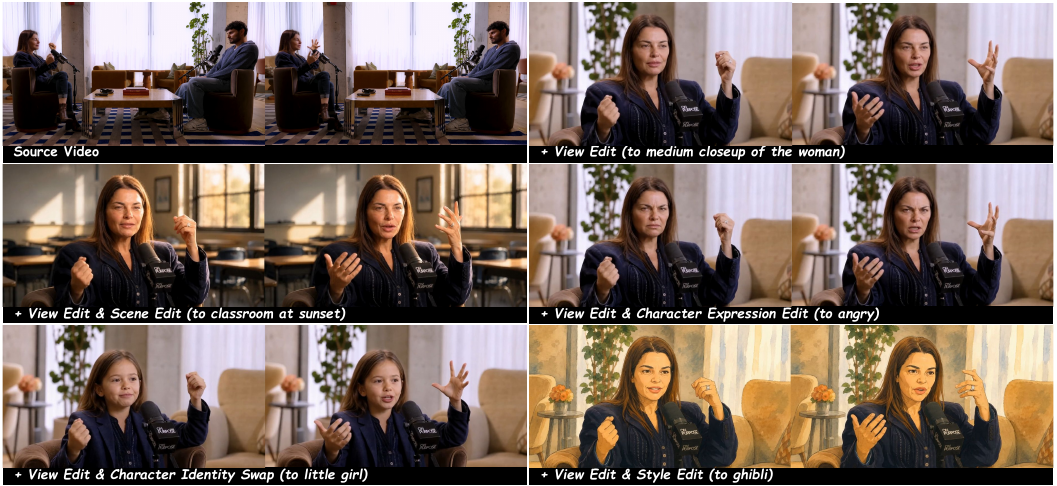}
  \caption{Post-hoc art direction. Given a single wide source video (top-left), our method synthesizes novel framings while propagating target-frame edits (scene, expression, identity, style) and preserving speech and motion.}
  \label{fig:application_edit}
\end{figure}

\paragraph{Applications.}
\label{sec:exp_applications}
Our method enables two applications: (1) Agentic editing: an LLM agent plans dialogue cuts from an editing-style prompt, such as ``cut on listener reactions,'' and calls our method to generate the shots it needs, editing a single take into a multi-shot scene end to end. (2) Post-hoc art direction: changing the scene, an actor's look, or the style normally means reshooting. Here, an edit to the target frame carries through the whole shot while the performance stays the same (Figures~\ref{fig:autocine_demo} and~\ref{fig:application_edit}).

\subsection{Ablation Study}
\label{sec:exp_ablation}
\label{sec:exp_gating_ablation}

\paragraph{Audio Conditioning and Speaker-Aware Gating.}
In Table~\ref{tab:ablation_audio}, we run inference with and without the audio input. Without audio, lip sync degrades on both benchmarks, while mouth LPIPS and MSE change only slightly, since the mouth still stays close to the target in appearance but moves out of sync. A generative model can produce realistic video from the visual conditions alone, but under large viewpoint shifts only the audio guidance can keep the lip motion in sync with the dialogue.

In Table~\ref{tab:gating_ablation}, we run inference with and without speaker-aware gating on $15$ in-the-wild crowd scenes. The gate nearly halves speech leakage and improves dialogue mAP. Without it, silent listeners tend to mouth the speaker's words (Figure~\ref{fig:ablation_gating}). Sync-C drops slightly because it averages over all visible faces~\cite{SyncNet2016}, so it penalizes silent faces, whose lips do not follow speech from another, possibly off-screen, speaker. On a speaker shot, gating leaves the lip sync visually unchanged because the gate is $1$ on active speakers.

\paragraph{Crop Augmentation.}
In Table~\ref{tab:ablation_training}, we train the model on the same data with and without crop augmentation and evaluate both variants with audio conditioning and gating enabled. Crop augmentation expands the range of framings at each camera viewpoint. Without it, the model sees a narrower range of target framings in training and handles new framings less robustly: facial, body, and dialogue consistency degrade, and some outputs even miss the target framing. Supplementary \suppref{Figure~\ref{fig:ablation_training}}{Section~D} shows qualitative examples.

\section{Conclusion, Limitations, and Future Work}
\label{sec:conclusion}

We presented a way to create multi-shot dialogue coverage from a single recording while preserving the original performance. Our starting point is that speech does not change with the camera. The original audio therefore provides a view-invariant cue for articulation, even when the speaker's lips are barely visible in the source. Speaker-aware audio gating carries speaking activity across views and reduces speech leakage. On paired views recorded by real cameras, our method more accurately reproduces the target shot and better preserves facial and body motion and dialogue activity than the baselines. An editor can thus cut close-ups, over-the-shoulder views, and listener reactions against the original audio, all from a single take, and explore how to tell the story after the performance is captured.

The quality of the new shot depends in part on the target frame and on correctly identifying who is speaking. Better target-frame generation could improve identity and pose consistency, and combining visual speaker detection with audio diarization could help resolve ambiguous cases. The video model also fills in details outside the source view with plausible estimates, so shots of real people should be made with their consent and disclosed as synthesized.

An intriguing direction is to explore whether other aspects of a performance can be described independently of the camera. A director-style description of gestures, gaze, and timing could provide a shared cue across shots, much as the original audio does for speech.

\bibliographystyle{ACM-Reference-Format}
\bibliography{references}

\setcounter{figure}{0}\setcounter{table}{0}\setcounter{equation}{0}\setcounter{algorithm}{0}
\renewcommand{\thefigure}{S\arabic{figure}}
\renewcommand{\thetable}{S\arabic{table}}
\renewcommand{\theequation}{S\arabic{equation}}
\renewcommand{\thealgorithm}{S\arabic{algorithm}}
\renewcommand{\theHfigure}{S\arabic{figure}}
\renewcommand{\theHtable}{S\arabic{table}}
\renewcommand{\theHequation}{S\arabic{equation}}
\renewcommand{\theHalgorithm}{S\arabic{algorithm}}
\appendix
\setlength{\emergencystretch}{3em}
\makeatletter
\setlength{\@fptop}{0pt}\setlength{\@fpsep}{14pt}\setlength{\@fpbot}{0pt plus 1fil}
\setlength{\@dblfptop}{0pt}\setlength{\@dblfpsep}{14pt}\setlength{\@dblfpbot}{0pt plus 1fil}
\makeatother

\section{Method Details}
\label{sec:app_impl}

\subsection{Training}
\label{sec:app_training}

We train on 9k pairs: 5k video-only pairs from CMU Panoptic~\cite{CMUPanoptic2017}, PKU-DyMVHumans~\cite{PKUDyMVHumans2024}, and SynCamVideo~\cite{SynCamVideo2025} and 4k audio-visual pairs from Replay~\cite{Replay2023}, MEAD~\cite{MEAD2020}, and DnD Group Gesture~\cite{ConvoFusion2024}, at $960{\times}768{\times}81$ resolution for $30$k steps on one H100 GPU. With the backbone frozen, rank-32 LoRA~\cite{LoRA2022} adapters are trained with learning rate $2\times10^{-4}$ on the video self-attention, text cross-attention, and feed-forward layers, and also on the audio-to-video cross-attention, which only the audio-visual pairs update. Source and target tokens share the same positional encodings. We train on the full checkpoint and run inference with the distilled eight-step checkpoint. Clips are $25$\,fps, and generating one takes $30$ seconds on one H100 GPU.

\begin{figure}[!htb]
  \centering
  \includegraphics[width=\columnwidth,pagebox=cropbox]{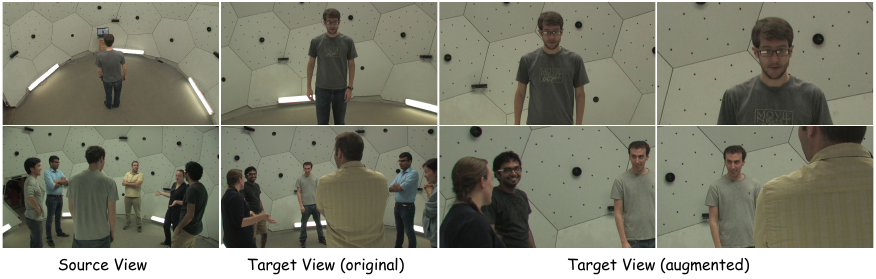}
  \caption{Training data construction. We estimate body poses on the original target videos and use them to crop the videos to increase framing diversity.}
  \Description{Two rows of four frames from a multi-camera capture dome. In each row, the first frame is the source view, the second is the original target view from another camera, and the last two are crops of the target view: a single person in a medium shot and a medium close-up in the first row, and a group in a medium three-shot and an over-the-shoulder shot in the second row.}
  \label{fig:data_construction}
\end{figure}

Figure~\ref{fig:data_construction} shows how training pairs are cropped to increase framing diversity (Section~\ref{sec:training}).

\subsection{Speaker-Aware Audio Gating}
\label{sec:app_gating}
\label{sec:appendix_gating}

Throughout, $t$ indexes latent video frames, $s$ denoising steps, $p$ source participants, $h$ the $H$ attention heads of width $d_h$, and $i$ and $j$ the spatial latent locations of the target and the source. Attention is captured in the first $s_g=2$ steps, $s\in\{0,1\}$.

\paragraph{Identity Matching.}%
\label{sec:app_gating_matching}

At each captured step $s$ and identity-matching block $\ell\in\mathcal{L}=\{32,33,34\}$, we take the video self-attention queries and keys of latent frame~$0$ of the target (the encoded $I_t$) and of the source, after RMS normalization~\cite{RMSNorm2019} and before rotary position encoding~\cite{RoFormer2024}, and average the scaled logits over heads in both directions:
\begin{equation}
\begin{aligned}
S^{(s,\ell)}_{ij}&=\frac{1}{H\sqrt{d_h}}\sum_{h=1}^{H}\big\langle Q^{\mathrm{tgt}}_{h,i},K^{\mathrm{src}}_{h,j}\big\rangle,\\
\bar S^{(s,\ell)}_{ji}&=\frac{1}{H\sqrt{d_h}}\sum_{h=1}^{H}\big\langle Q^{\mathrm{src}}_{h,j},K^{\mathrm{tgt}}_{h,i}\big\rangle.
\end{aligned}
\label{eq:gating_correspondence}
\end{equation}
We normalize each over its key locations, multiply the two directions, and take the square root:
\begin{equation}
W_{ct}(i,j)=\frac{1}{2|\mathcal{L}|}\sum_{s=0}^{1}\sum_{\ell\in\mathcal{L}}
\sqrt{\operatorname*{softmax}_{j}\big(S^{(s,\ell)}_{ij}\big)\,\operatorname*{softmax}_{i}\big(\bar S^{(s,\ell)}_{ji}\big)}.
\label{eq:gating_weights}
\end{equation}
Rows index target locations and columns index source locations.

\paragraph{Face Tracking.}%
\label{sec:app_gating_masks}
Following DiffTrack~\cite{Nam2025DiffTrack}, we track each located face with the bidirectional attention between the clean target frame and each later latent frame at block~$33$, averaged over heads and the two captured steps. Each facial patch is placed at its best-matching location in frame~$t$, and the face center $c_{p,t}$ follows the mean displacement of the patches from their positions in the target frame, smoothed over time. The mask $M_{p,t}$ is an axis-aligned Gaussian centered on $c_{p,t}$, with a spread set by the extent of the face in the latent grid.

\paragraph{Gate Schedule.}
\label{sec:app_gating_schedule}

All $48$ audio-to-video blocks are set to zero during the first $s_g$ steps and use the gate of Equation~\ref{eq:speaker_gate} in the remaining steps.
Algorithm~\ref{alg:gating} gives the full procedure.

\begin{algorithm}[!ht]
  \caption{Speaker-aware audio gating at inference.}
  \label{alg:gating}
  \small
  \begin{algorithmic}[1]
    \REQUIRE source video $V_s$, audio $A$, target frame $I_t$
    \ENSURE generated target video $\hat V_t$
    \STATE $y_{p,t}\leftarrow$ LightASD$(V_s,A)$, thresholded and resampled to latent time
    \STATE Map source facial patches $\mathcal{S}_{p,a}$ at the first source frame onto the latent grid
    \STATE Initialize target tokens: clean $I_t$ latent at frame $0$, noise elsewhere
    \FOR{$s=0,1$}
      \STATE Denoise with all A2V updates set to zero
      \STATE Capture $Q,K$ from visual self-attention in blocks $32$--$34$
    \ENDFOR
    \STATE Compute $W_{ct}$ (Equations~\ref{eq:gating_correspondence}--\ref{eq:gating_weights})
    \STATE Locate each participant's face in $I_t$ (Equation~\ref{eq:face_localization}) and discard faces with weak or ambiguous correspondence
    \FOR{each located face $p$}
      \STATE Track $c_{p,t}$ with block-$33$ attention
      \STATE Build mask $M_{p,t}$
    \ENDFOR
    \STATE Build $G$ (Equation~\ref{eq:speaker_gate}) and freeze it
    \FOR{$s=2,\ldots,7$}
      \STATE Denoise, with every A2V block outputting $G_t(i)\,\Delta h_t(i)$
    \ENDFOR
    \STATE \textbf{return} $\hat V_t\leftarrow$ decoded target tokens
  \end{algorithmic}
\end{algorithm}

\section{Experimental Details}
\label{sec:app_eval}

\subsection{Benchmarks}
\label{sec:supp_bench_stats}

\begin{figure*}[t]
  \centering
  \includegraphics[width=\textwidth]{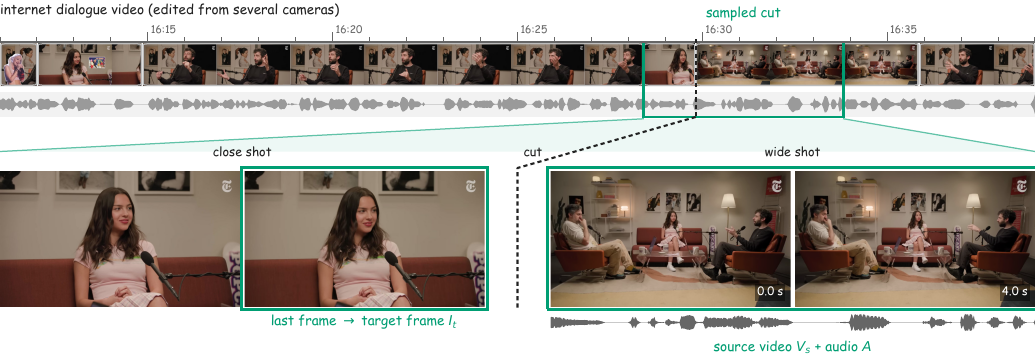}
  \caption{In-the-wild pair construction. At a cut from a close shot to a wide shot, the wide shot and its audio form the source video $V_s$, and the last frame of the close shot is the target frame $I_t$.}
  \label{fig:supp_inthewild_construction}
\end{figure*}

Figure~\ref{fig:supp_inthewild_construction} shows how an in-the-wild pair is built. Each internet video is edited by its creators from several cameras and cuts from a close shot to a wide shot of the same conversation. The wide shot, with its audio, serves as the source video $V_s$, and the last frame of the close shot serves as the target frame $I_t$. The conversation continues across the cut, so this frame shows the people at the moment the wide shot begins, as the target frame should (Section~\ref{sec:cinematic_control}). Each person in $V_s$ is labeled as a speaker or a listener from the reference speaking labels (Section~\ref{sec:supp_metrics}), and the faces in $I_t$ are matched to these people by face-embedding similarity.

\subsection{Evaluation Metrics}
\label{sec:supp_metrics}

\paragraph{Appearance.}
We measure perceptual distance with LPIPS~\cite{LPIPS2018} and pixel error with the mean squared RGB error (MSE). Both are computed on the full frame against the ground-truth target view. Identity similarity is the cosine similarity between ArcFace~\cite{ArcFace2019} embeddings of the generated and ground-truth faces.

\paragraph{Facial Performance.}
Landmark distance (LMD)~\cite{LMD_Chen2018} measures the mean distance between facial landmarks of the generated and reference faces, normalized by the reference face size. Facial expression error (AED) is the mean absolute difference between DECA~\cite{DECA2021} expression coefficients. For gaze error, we compute the angle between head-relative gaze directions estimated with OpenFace~2~\cite{OpenFace2_2018}.

\paragraph{Body Performance.}
\looseness=-1 We estimate $12$ body joints with RTMPose~\cite{RTMPose2023} and keep joints with a confidence of at least $0.5$. PCK~\cite{PCK_YangRamanan2013} is the percentage of generated joints that lie within $0.05\sqrt{a_{\mathrm{box}}}$ of the reference joint, where $a_{\mathrm{box}}$ is the area of the reference person box. Velocity error compares joint velocities between consecutive frames, normalized by the reference body scale.%

\paragraph{Speaker Attribution.}
We use TalkNet~\cite{TalkNet2021} to obtain reference speaking labels and per-face scores on the output. Dialogue mAP is the mean of the average precision for speaking and for non-speaking activity. Speech leakage is the share of frames in which a face that is silent in the source speaks in the target output. We measure lip synchronization with SyncNet~\cite{SyncNet2016} confidence (Sync-C) and distance (Sync-D), averaged over all visible faces.

\paragraph{View Following.}
A model can keep identity similarity high by copying the source video instead of moving to the target view. Source copy is the percentage of clips whose output, at the majority of sampled timestamps, is closer in SSIM~\cite{SSIM2004} to the source view than to the ground-truth target view.

\subsection{Baseline Implementation}
\label{sec:supp_baselines}
Every baseline receives the source video and the target frame of each clip, the source audio when it accepts audio, and a clip-independent text prompt. %
We run the open-source MiniMax~H3~\cite{MiniMaxH3_2026} with its default settings ($12$ denoising steps), $4$-second outputs, and a fixed seed, and the Wan~3.0~\cite{Wan3_2026} API at $720$p with prompt expansion enabled.
 The camera-controlled models GEN3C~\cite{GEN3C2025} and Vista4D~\cite{Vista4D2026} take a target camera pose instead of the target frame. We estimate the pose by reconstructing the source video with $\pi^3$~\cite{Pi3_2025} and registering the target frame to it with RoMa~v2~\cite{RoMaV2_2025} matches and PnP. Neither takes audio, so the source audio is attached to their outputs. The reference-guided video editors VINO~\cite{VINO2026} and Bernini-R~\cite{Bernini2026} take the source video and the target frame but no audio.

\section{Additional Experiments}
\label{sec:app_results}

\subsection{Target-Frame Generator}
\label{sec:app_first_frame}
\begin{figure*}[t]
  \centering
  \begin{minipage}{\textwidth}%
    \includegraphics[width=\linewidth,pagebox=cropbox]{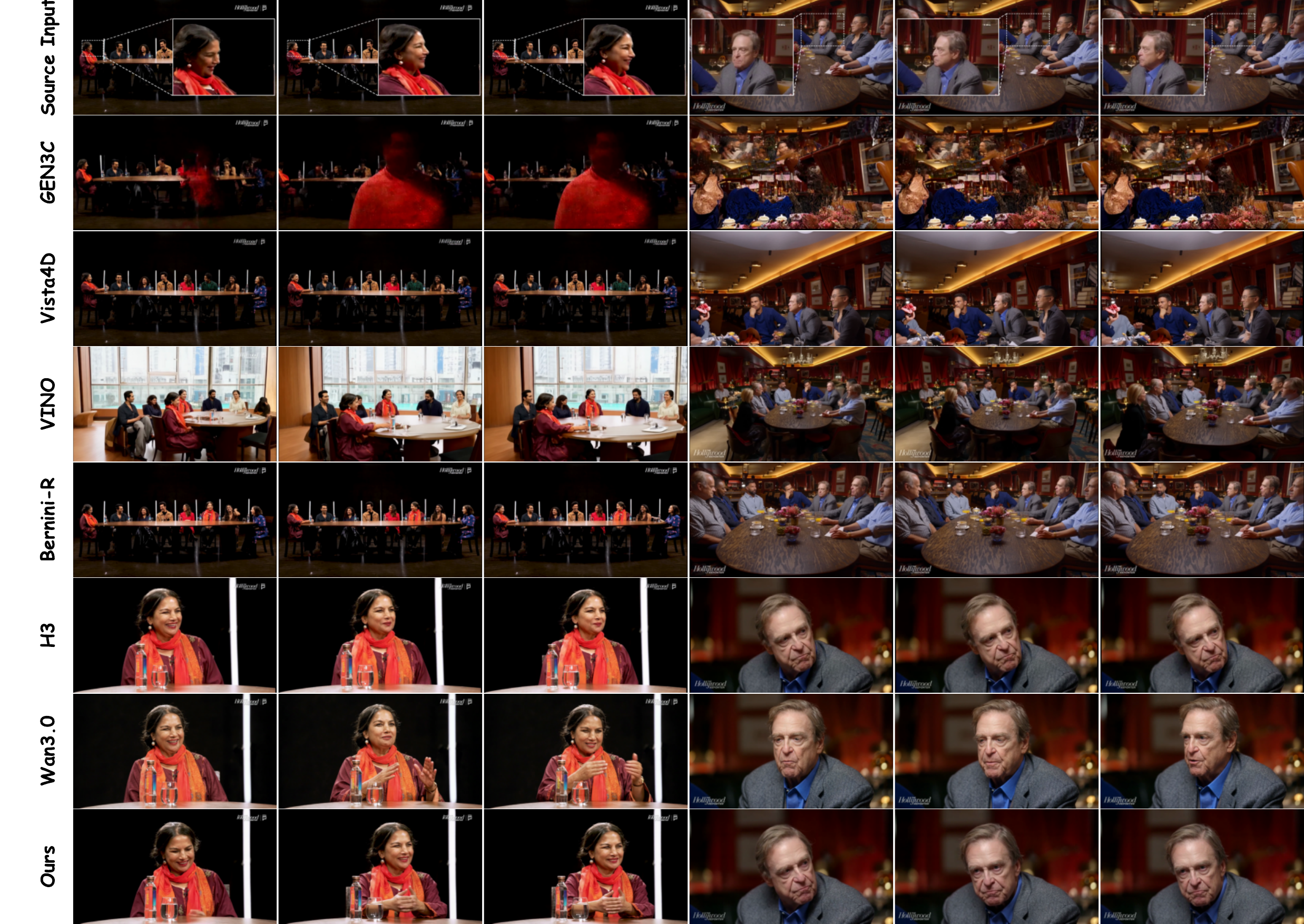}%
  \caption{Qualitative comparison on the in-the-wild benchmark, including the camera-controlled models GEN3C~\cite{GEN3C2025} and Vista4D~\cite{Vista4D2026} and the reference-guided video editors VINO~\cite{VINO2026} and Bernini-R~\cite{Bernini2026} (implementation in Section~\ref{sec:supp_baselines}). Only H3, Wan~3.0, and our model reach the requested close-up, and ours also follows the source motion and speech.}
  \Description{Two examples side by side, each with eight rows of three frames. The top row shows the wide source input with a dashed box around the target subject and an enlarged inset of that subject; the following rows show results from GEN3C, Vista4D, VINO, Bernini-R, H3, Wan 3.0, and ours. The left example is a woman in an orange scarf at a long panel table; the right example is an older man in a grey suit at a round table in a wood-paneled room. GEN3C shows heavy artifacts, Vista4D, VINO, and Bernini-R show wide views of the table, and H3, Wan 3.0, and ours show close-ups of the subject.}
  \label{fig:inthewild_camctrl}
  \end{minipage}
\end{figure*}
\begin{figure*}[t]
  \centering
  \includegraphics[width=0.90\textwidth]{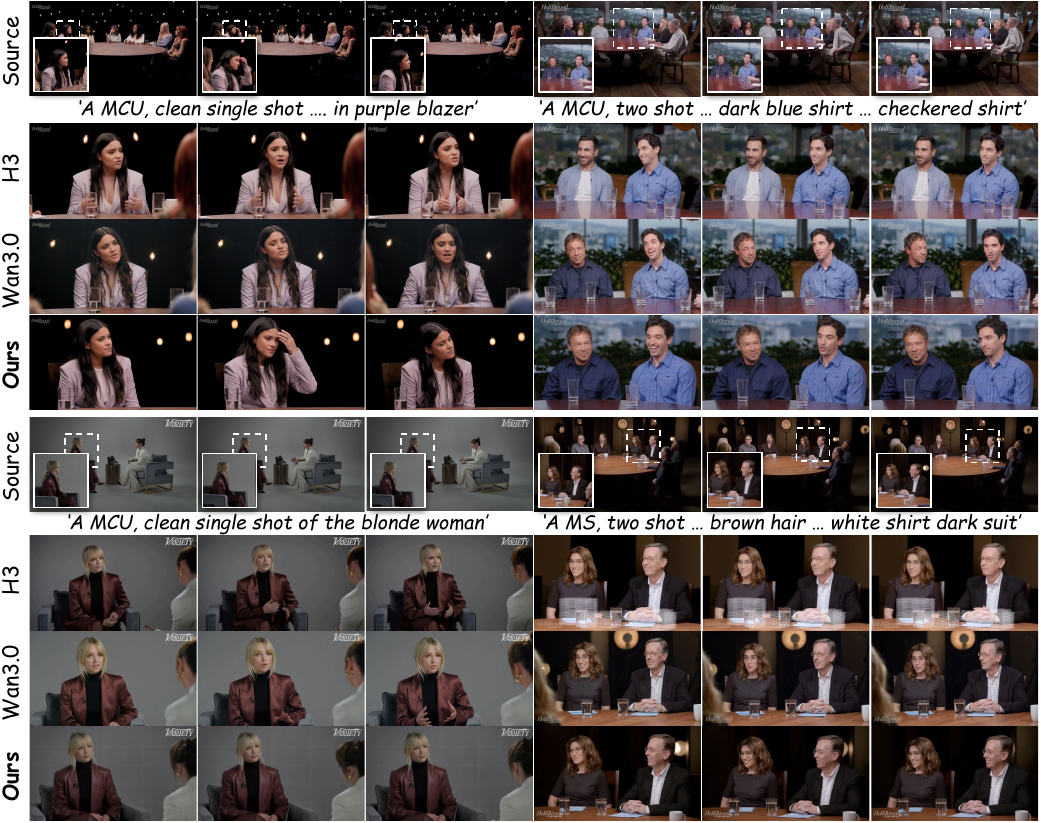}
  \caption{End-to-end comparison from a shot prompt. Four examples compare MiniMax~H3, Wan~3.0, and our full pipeline. Each example shows the source take and requested shot above the generated frames.}
  \Description{Four examples arranged in two rows and two columns. Each example contains a source sequence, shot prompt, and results from H3, Wan 3.0, and ours.}
  \label{fig:qual_comp}
\end{figure*}
\begin{figure}[t]
  \centering
  \includegraphics[width=\columnwidth,pagebox=cropbox]{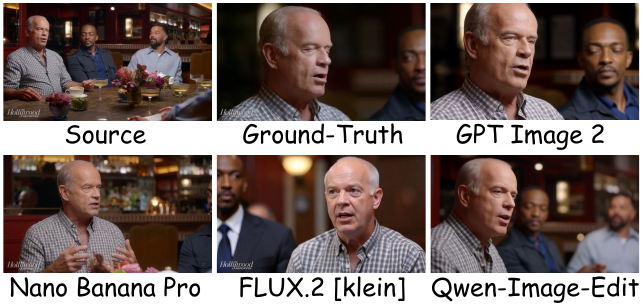}
  \caption{Target frames generated by each editor from the source frame and a shot prompt, with the ground-truth target view. GPT-Image~2 comes closest to the ground-truth framing.}
  \Description{Two-row qualitative grid. Top row: source frame, ground-truth target view, and the GPT Image 2 result. Bottom row: results from Nano Banana Pro, FLUX.2 [klein], and Qwen-Image-Edit. The GPT Image 2 result most closely matches the ground-truth over-the-shoulder framing.}
  \label{fig:qual_image}
\end{figure}

An image editor synthesizes the target frame $I_t$ from the source video's first frame and a shot prompt (Section~\ref{sec:cinematic_control}). Table~\ref{tab:image_generation_benchmark} compares such editors on clips from the in-the-wild benchmark. GPT-Image~2 performs best, leading on background similarity and shot type, so we use it for the target frames in Figures~\ref{fig:qual_shots_more}, \ref{fig:qual_comp}, and~\ref{fig:qual_shots_more_supp}. Figure~\ref{fig:qual_image} shows the editors' outputs on one clip.

\begin{table}[!ht]
  \centering
  \caption{Target-frame generators. DINO~\cite{DINO2021}: background similarity. Size/Type: pass rates (\%) for shot size and type, where a frame passes if the body pose detected in it meets the pose criteria of the requested shot.}
  \label{tab:image_generation_benchmark}
  \footnotesize
  \setlength{\tabcolsep}{1.5pt}
  \renewcommand{\arraystretch}{1.1}
  \begin{tabular*}{\columnwidth}{@{\extracolsep{\fill}}lccc@{}}
    \toprule
    Model & DINO\,$\uparrow$ & Size\,$\uparrow$ & Type\,$\uparrow$ \\
    \midrule
    HunyuanImage 3.0~\cite{HunyuanImage3_2025} & 0.592 & 50.0 & 54.5 \\
    FLUX.2 [klein] 4B~\cite{FLUX2Klein2026} & 0.620 & \underline{72.7} & 61.4 \\
    Qwen-Image-Edit~\cite{QwenImage2025} & \underline{0.632} & \textbf{75.0} & 61.4 \\
    Nano Banana Pro~\cite{NanoBananaPro2025} & 0.596 & 65.9 & \underline{63.6} \\
    GPT-Image 2~\cite{GPTImage2_2026} (selected) & \textbf{0.679} & \underline{72.7} & \textbf{70.5} \\
    \bottomrule
  \end{tabular*}
\end{table}
\begin{figure}[t]
  \centering
  \includegraphics[width=0.74792\columnwidth,pagebox=cropbox,trim=0 0 337.4 0,clip]{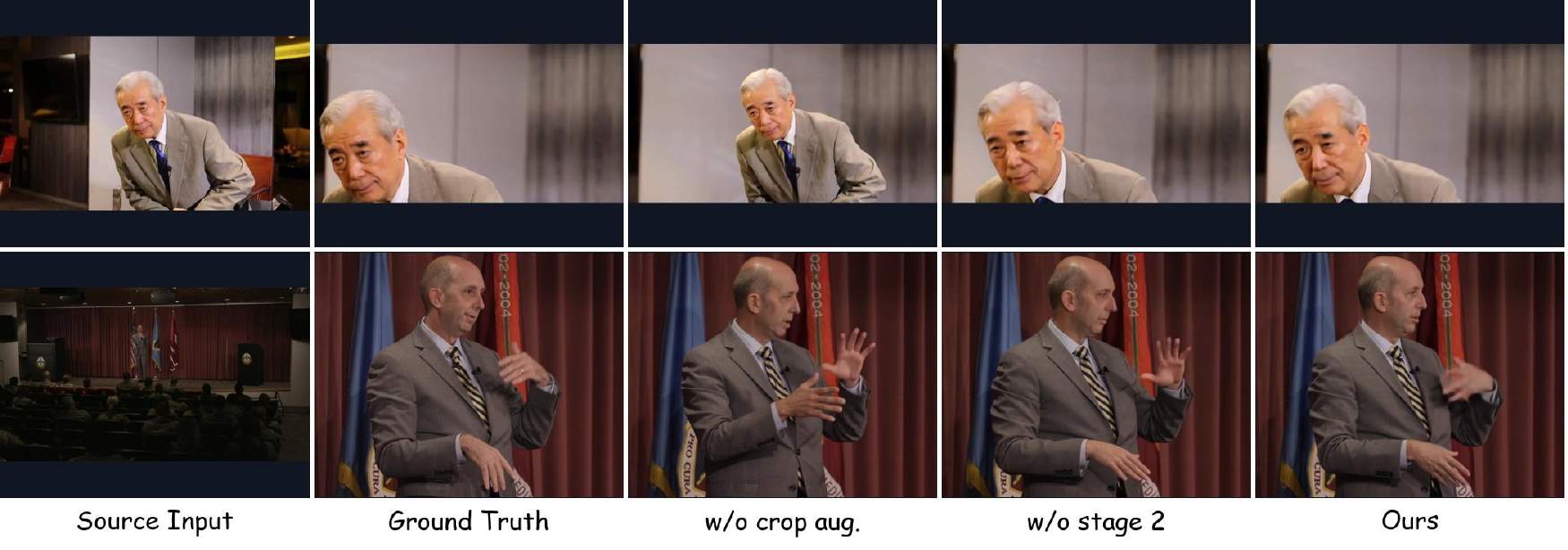}%
  \includegraphics[width=0.25208\columnwidth,pagebox=cropbox,trim=670.6 0 0 0,clip]{figures/ablation_train.pdf}
  \caption{Qualitative ablation on crop augmentation. Without crop augmentation, the output misses the target framing (top), and the gesture departs from the ground truth (bottom).}
  \Description{Two rows of four frames. Columns: source input, ground truth, without crop augmentation, and ours. Top row: an older man in a suit leaning forward; the result without crop augmentation is framed wider than the ground truth. Bottom row: a speaker at a podium event, shown in a medium shot; the result without crop augmentation raises the hands differently from the ground truth, while ours follows the recorded gesture more closely.}
  \label{fig:ablation_training}
\end{figure}

\section{Additional Visualizations}
\label{sec:app_vis}
\begin{figure*}[p]
  \centering
  \includegraphics[width=\textwidth]{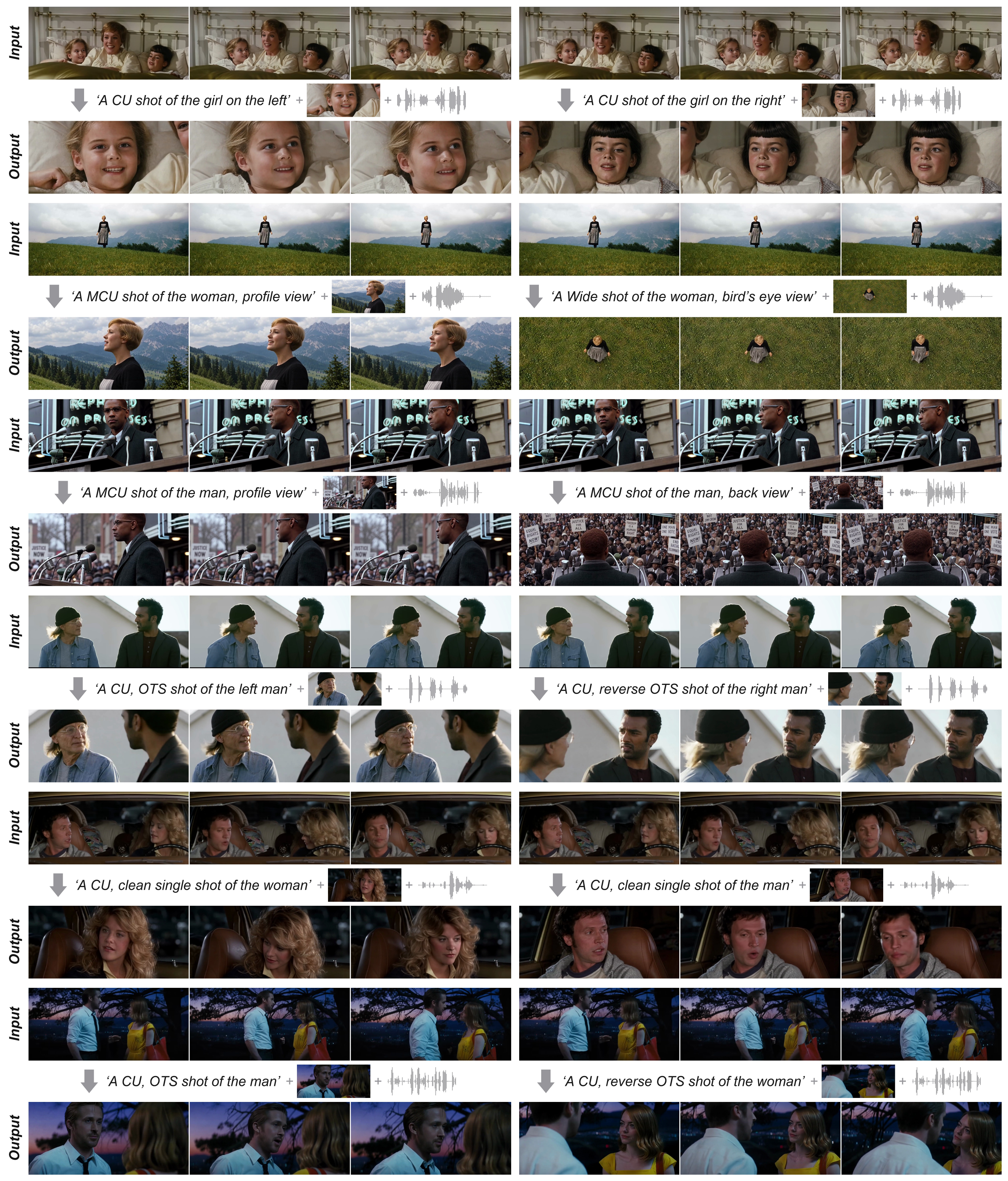}
  \caption{Results on six more clips. Target frames are generated by GPT-Image~2 from the source frame and the shot prompt.}
  \label{fig:qual_shots_more_supp}
\end{figure*}

Figure~\ref{fig:qual_shots_more_supp} shows results on six more clips, with target frames generated from shot prompts.
Figure~\ref{fig:ablation_training} shows the qualitative ablation on crop augmentation. %

\newcommand{\suppInTheWildQual}{%
\subsection{In-the-Wild Qualitative Comparison}
\label{sec:supp_results}
Figure~\ref{fig:inthewild_camctrl} compares our method with the baselines, camera-controlled models, and reference-guided video editors on two in-the-wild clips. The camera-controlled models and the reference-guided video editors struggle with the large change of viewpoint. GEN3C breaks down into artifacts, and Vista4D, VINO, and Bernini-R stay at the wide source framing. The reference-guided video generation models H3~\cite{MiniMaxH3_2026} and Wan~3.0~\cite{Wan3_2026} reach the requested close-up and move the lips with the speech. However, they generate their own gestures and expressions instead of following the source performance, and they can make silent listeners mouth the speaker's words.
}
\suppInTheWildQual

\subsection{End-to-End Qualitative Comparison}
\label{sec:supp_prompt_qual}
Figure~\ref{fig:qual_comp} compares MiniMax~H3, Wan~3.0, and our method end to end on in-the-wild clips, where the new shot is requested with a shot prompt instead of a target frame. Our pipeline first generates the target frame $I_t$ from the prompt with GPT-Image~2 (Section~\ref{sec:cinematic_control}), and H3 and Wan~3.0 receive the source video, the audio, and the shot prompt. The examples show three baseline failure modes.
First, speech is attributed to the wrong face: both baselines animate the listener's mouth in the two left examples, as does Wan~3.0 in the lower-right. In the upper-right two-shot, H3 silences the right-hand man while he speaks, and Wan~3.0 keeps him speaking after he falls silent.
Second, the generated motion deviates from the captured performance: both baselines produce hand gestures that do not match the source in the two left examples and, in the lower-left, direct the subject's gaze toward the opposite participant.
Third, framing and appearance drift: H3 pulls the wrong participant into the upper-right two-shot and leaves ghosted outlines of glasses and hands in the lower-right.

\end{document}